\documentclass[letterpaper]{article} % DO NOT CHANGE THIS
\usepackage{aaai23}  % DO NOT CHANGE THIS
\usepackage{times}  % DO NOT CHANGE THIS
\usepackage{helvet}  % DO NOT CHANGE THIS
\usepackage{courier}  % DO NOT CHANGE THIS
\usepackage[hyphens]{url}  % DO NOT CHANGE THIS
\usepackage{graphicx} % DO NOT CHANGE THIS
\usepackage{natbib}  % DO NOT CHANGE THIS AND DO NOT ADD ANY OPTIONS TO IT
\usepackage{caption} % DO NOT CHANGE THIS AND DO NOT ADD ANY OPTIONS TO IT
\usepackage{url}
\usepackage{graphicx}
\usepackage{amsmath}
\usepackage{booktabs}
\usepackage{bbm}
\usepackage{graphicx}
\usepackage{newfloat}
\usepackage{listings}
\usepackage{color}
\usepackage{amssymb}
\usepackage{multirow}
\usepackage{multicol}
\usepackage{lineno,hyperref}
\modulolinenumbers[5]

\usepackage{amsmath,amssymb,amsfonts}
\usepackage{algorithmic}
\usepackage{algorithm}
\usepackage{graphicx}
\usepackage{textcomp}
\usepackage{colortbl}
\usepackage{xcolor}
\usepackage{color}
\usepackage{listings}
\usepackage{xcolor}
\usepackage{algorithm}
\usepackage{algorithmic}
\makeatletter
\newcommand*\bigcdot{\mathpalette\bigcdot@{.5}}
\newcommand*\bigcdot@[2]{\mathbin{\vcenter{\hbox{\scalebox{#2}{$\m@th#1\bullet$}}}}}
\makeatother
\usepackage{newfloat}
\usepackage{listings}
\DeclareCaptionStyle{ruled}{labelfont=normalfont,labelsep=colon,strut=off} % DO NOT CHANGE THIS
\floatstyle{ruled}
\newfloat{listing}{tb}{lst}{}
\floatname{listing}{Listing}
\usepackage{bibentry}
\begin{document}
\title{TQ-Net: Mixed Contrastive Representation Learning For Heterogeneous Test Questions}
\author{He Zhu$^{1,3}$, Xihua Li$^{2}$, Xuemin Zhao$^{2}$, Yunbo Cao$^{2}$, Shan Yu$^{1,3}$}
\affiliations{$^1$Brainnetome Center, National Laboratory of Pattern Recognition (NLPR), CASIA\\
$^2$ Tencent Inc. China \\
		$^3$School of Future Technology, University of Chinese Academy of Sciences (UCAS)\\
	 he.zhu@nlpr.ia.ac.cn, lixihua9@126.com, xueminzhao@tencent.com\\
	 yunbocao@tencent.com, shan.yu@nlpr.ia.ac.cn\\
}
\maketitle
\begin{abstract}
Recently, more and more people study online for the convenience of access to massive learning materials (e.g. test questions/notes), thus accurately understanding learning materials became a crucial issue, which is essential for many educational applications. Previous studies focus on using language models to represent the question data. However, test questions (TQ) are usually heterogeneous and multi-modal, e.g., some of them may only contain text, while others half contain images with information beyond their literal description. In this context, both supervised and unsupervised methods are difficult to learn a fused representation of questions. Meanwhile, this problem cannot be solved by conventional methods such as image caption, as the images may contain information complementary rather than duplicate to the text. In this paper, we first improve previous text-only representation with a two-stage unsupervised instance level contrastive based pre-training method (\textbf{MCL}: \textbf{M}ixture Unsupervised \textbf{C}ontrastive \textbf{L}earning). Then, \textbf{TQ-Net} was proposed to fuse the content of images to the representation of heterogeneous data. Finally, supervised contrastive learning was conducted on relevance prediction-related downstream tasks, which help the model to effectively learn the representation of questions. We conducted extensive experiments on question-based tasks on large-scale, real-world datasets, which demonstrated the effectiveness of TQ-Net and improve the precision of downstream applications (e.g. similar questions $\uparrow$2.02\% and knowledge point prediction $\uparrow$7.20\%). Our code will be available, and we will open-source a subset of our data to promote the development of relative studies. 
\end{abstract}

\section{Introduction}
\par Nowadays, more and more users prefer to learn courses or take tests from the online learning system. Owning a great number of question materials, these platforms are expected to design an automatic model to provide personalized test and practice lists to effectively improve the weak knowledge of different users, such as recommending similar or related knowledge-point exercises to users, which they are poorly learned.
\begin{figure}[h]
\centering
    \includegraphics[width=0.40\textwidth, trim=0 0 0 0, clip]{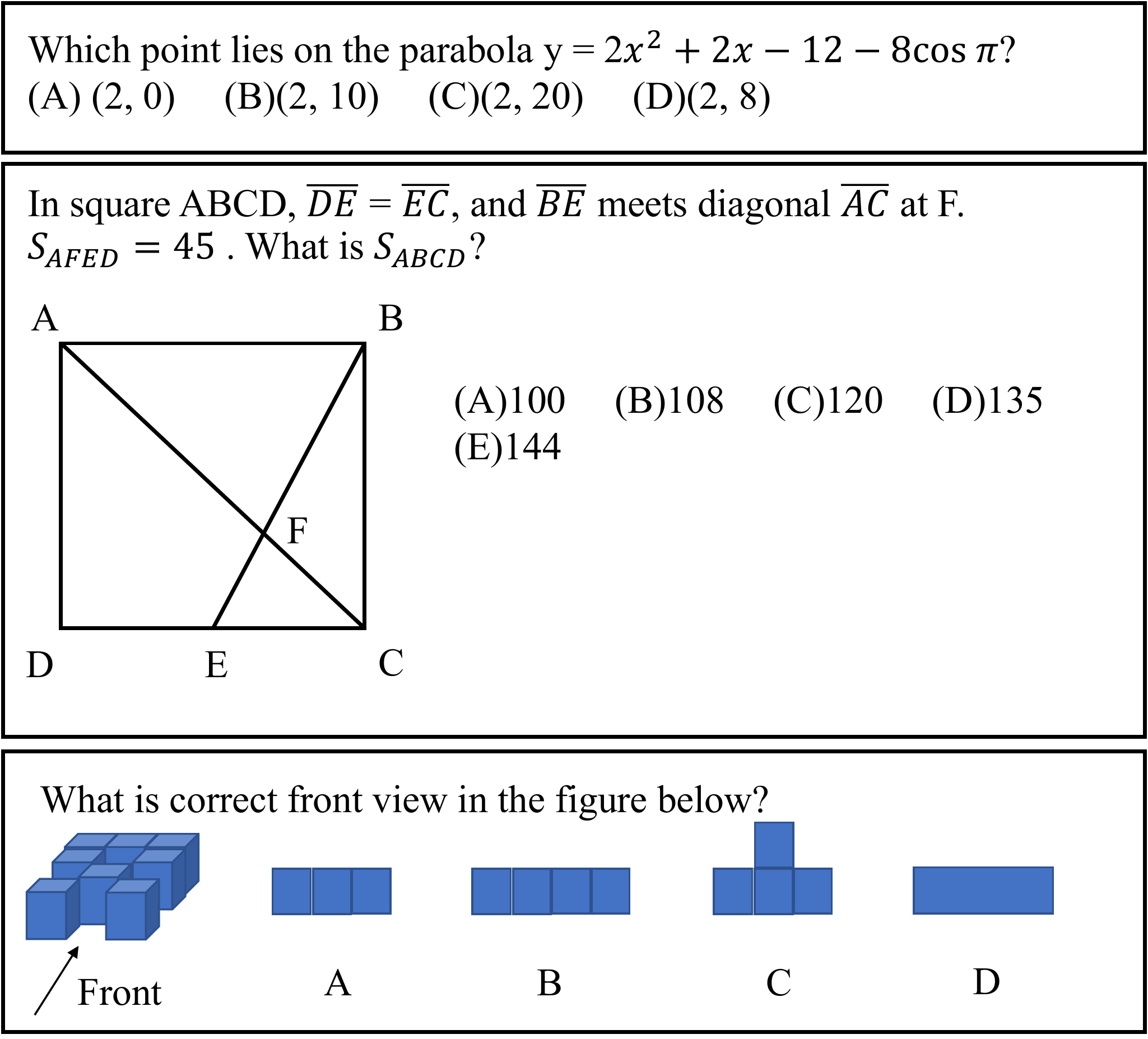}
% \begin{minipage}[t]{1\linewidth}
% 	\centering
% 	    \includegraphics[width=1\textwidth, trim=0 0 0 0, clip]{q1.pdf}
% 	\centerline{(a) Question with formula and text.}
% \end{minipage} 
% \begin{minipage}[t]{1\linewidth}
% 	\centering
% 	    \includegraphics[width=0.8\textwidth, trim=0 0 0 0, clip]{q2.pdf}
% 	\centerline{(b) Question with formula and image input.}\medskip
% \end{minipage} 
% \begin{minipage}[t]{1\linewidth}
% 	\centering
% 	    \includegraphics[width=0.8\textwidth, trim=0 0 0 0, clip]{q3.pdf}
% 	\centerline{(c) Question with multiple images input.}\medskip
% \end{minipage} 
    \caption{Examples of heterogeneous questions. The heterogeneous question samples have various data structure, contains formula, text and images. Moreover, the information of text, formula and image may have different importance in various questions.}
    \label{fig:q_data}
\end{figure}
\par The key ability of this automatic model would be distinguishing various question data, or we can say, extracting the better representation of question data. In previous works, language models (LM) \cite{yu2014similarity,huang2017question,pardos2017imputing,anuyah2019using} were the main solutions: models trained with the supervised questions text data. However, language models cannot satisfy all subjects, like geometry questions for example. In other words, images contain information that are also important to lots of situations. 
\begin{figure*}[h]
    \centering
    \includegraphics[width=0.8\textwidth, trim=0 0 0 0, clip]{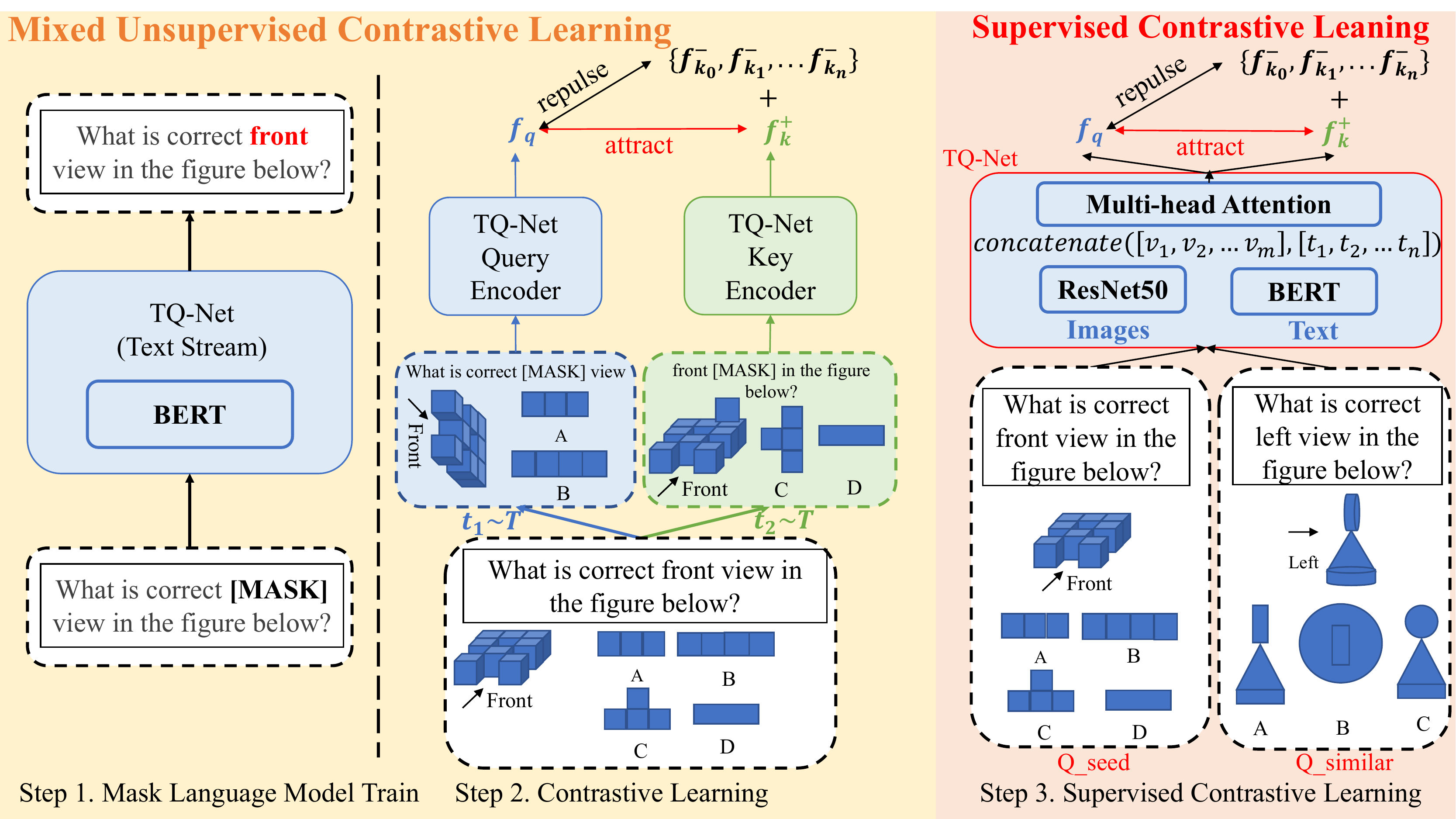}
    \caption{Training pipeline and TQ-Net model architecture. TQ-Net model simply consist of ResNet(visual encoder), BERT(text encoder) and multi-head attention(fusion module), visual tokens features come from multiple images will be concatenated with text token features and fusd by the attention module, at last output average pooling result as the representation of the input. Our framework has three steps: Step 1: The text encoder (BERT) will be self-supervised pre-trained by the mask language model task; Step 2: The multi-modal encoder will be pre-trained by CL with two separate data augmentation operators sampled from the same family of augmentations ($t_1\sim T$ and $t_2\sim T$) and applied to each question instance to obtain two correlated views. Step 3: The multi-modal encoder will be fine-tuned by choosing the positive and negative pairs based on the human-labelled similar questions to do the supervised contrastive learning.}
    \label{fig:pipeline}
\end{figure*} 
\par But the representation of questions cannot be solved by conventional methods such as image caption,  as the images may contain information complementary to the text. It should explore a balance between language model (LM) and multi-modal methods. How to design an effective representation learning method of the heterogeneous questions became a hard and valuable problem. 

\par Moreover, supervised methods pay expensively for the detailed labels marked by the different professionals. Recent unsupervised pre-training studies of LM or multi-modal \cite{devlin2018bert,sun2019videobert,li2020unicoder,su2019vl,xie2018rethinking,sun2019learning,tan2019lxmert,zhou2020unified} show that self-supervised-learning (SSL) was potential to be close to or even better than the representation learning capability of supervised methods without expensive cost.
\par From view of that, we design the TQ-Net to deal with the heterogeneous input and propose a contrastive learning (CL) pre-training\cite{chen2020simple,he2020momentum,grill2020bootstrap,chen2020exploring,wu2020clear,radford2021learning,zbontar2021barlow} method to solve the dilemma of representation learning. 
\par On the basis of all the above, we proposed two strategies to achieve a better unsupervised pre-training for questions: (1) Separately pre-training the image/text stream encoders and fine-tuning together. (2)Uniformly pre-training the two-stream encoder. Experiments show that the latter was more beneficial to the representation learning of questions.
\par Our contribution could be concluded as follows:
\begin{itemize}
    \item We proposed the \textbf{MCL} (\textbf{M}ixture Unsupervised \textbf{C}ontrastive \textbf{L}earning) mechanism to improve the text representation of questions. MCL is a two-stage unsupervised pre-training method which means that the model is first pre-trained in MLM (Masked Language Model) and then pre-trained in instance level CL (Contrastive Learning) framework (vs. modal level). The performance of MCL on both frozen and fine-tuning experiments are much better than the masked LM method.
    \item We proposed a new model named \textbf{TQ-Net} and instance level CL conducted to fuse the content of images to the representation of heterogeneous data. Compared to previous question-based model (text only ones and multi-modal ones), our proposal improves precision of downstream applications significantly.
    \item We proposed a supervised contrastive learning framework, named \textbf{SCL}, which help the model to effectively learn the representation of questions on relevance-related downstream task than traditional methods.
\end{itemize}
% \begin{figure*}[h]
%     \centering
%     \includegraphics[width=1.0\textwidth, trim=0 0 0 0, clip]{msbn.pdf}
%     \caption{Multi-modal shuffle BN before fusion. Our contrastive learning method is similar to the MoCo framework, but the differences are we use multi-modal shuffle BN before fusion to the attention module as shown in (b).}
%     \label{fig:msbn}
% \end{figure*}
\section{Related Work}
\par \noindent \textbf{Question Representation Learning} \cite{yu2014similarity} defined the representation and similarity measure of text questions based on a word and knowledge tree strategy. \cite{pardos2017imputing} used the skip-grams and bag-of-words to formulate the representation. \cite{liu2018finding} designed the MERL: a model consists of CNN and LSTM module, and fused the embeddings of various modals, learning the representation by the supervised training. \cite{yin2019quesnet} proposed the QuesNet: a Holed-Language-Model to extract the representation of questions data. This method needed three steps: (1) Embedding pre-training by auto-encoder for recovering the modal data to learn the embedding of different modal inputs. (2) Model pre-training for filling up every word with both the left and right side context of it. (3) Fine-tuning the whole network on the downstream tasks. QuesNet was complex to deploy and the performance would be limited because the embedding extractor and the model were separately trained.\\ 
\par \noindent \textbf{Unsupervised Pre-training of Multi-modal Content} VideoBERT \cite{sun2019videobert} built the dictionary for videos, and proposed using visual and text tokens as the input of the model. This method borrowed the idea of BERT \cite{devlin2018bert} to do the pre-training task: (1) Masked token prediction (2) Text-Video matching prediction. \cite{li2020unicoder,su2019vl,zhou2020unified} applied images as the visual input, these works introduced a detector or improved the pre-training tasks to capture the semantic information for better fusion. Moreover, CLIP \cite{2021Learning} aims to maximize the similarity between different modal inputs, but it is hard to take advantage of the difference in various modals to build a cooperative representation learning of heterogeneous data.\\
\par \noindent \textbf{Contrastive learning.} The CL compares two augmented views of the same input to capture semantic-invariant features, such as object-related characteristics. Some studies \cite{Chen2020A,He2019Momentum,chen2020big,Chen2020Improved,DBLP:conf/iccv/ChenXH21} attract the positive (similar) pairs and repulse negative (different) pairs to learn the representations. Other studies \cite{chen2020exploring,caron2021emerging} also provide a self-distillation framework, which only matches positive samples to learn the representations.
% \par \noindent \textbf{Contrastive Learning} CL was a great powerful SSL framework to pre-train models without any human knowledge. Unlike the pre-training task of BERT/VideoBERT which only could deal with the discrete signals, the CL methods were capable of learning the latent representation of continuous input such as images. The main idea of CL was instance discrimination, and its key was to prevent the model from collapsing during pre-training. The main community of CL was computer vision, but there were some significant CL works in the NLP \cite{wu2020clear} and multi-modal \cite{radford2021learning} area. Note that \cite{radford2021learning} was a multi-modal CL framework, but it still followed the idea of unsupervised multi-modal pre-training, which maximized the similarity between different modal of input. Thus, this method was hard to apply to heterogeneous question data.\\
\section{Method}
\subsection{Network Architecture}
\par We first introduce the novel two-forward model TQ-Net as shown in Figure \ref{fig:pipeline} and \ref{fig:fuse} to effectively extract the representation of the heterogeneous test data. Our proposal uses the text encoder and image encoder to respectively process the words and images input. 
%\subsection{Network Architecture}
% \begin{figure}[htbp]
%     \centering
%     \includegraphics[width=0.45\textwidth, trim=0 0 0 0, clip]{attention.pdf}
%     \caption{Self attention module architecture. }
%     \label{fig:attention}
% \end{figure}

\par Note that, here we use the pre-norm structure transformer to build self attention module as shown in Figure \ref{fig:fuse} (b). And the shortcut was introduced to the forward pass, which used to fuse the original modal information, it can also be interpreted as fusion of joint (low-level) and coordinated (high-level) information without lose of details.
\begin{figure*}[tbp]
 \centering
   \includegraphics[width=0.8\textwidth, trim=0 0 0 0, clip]{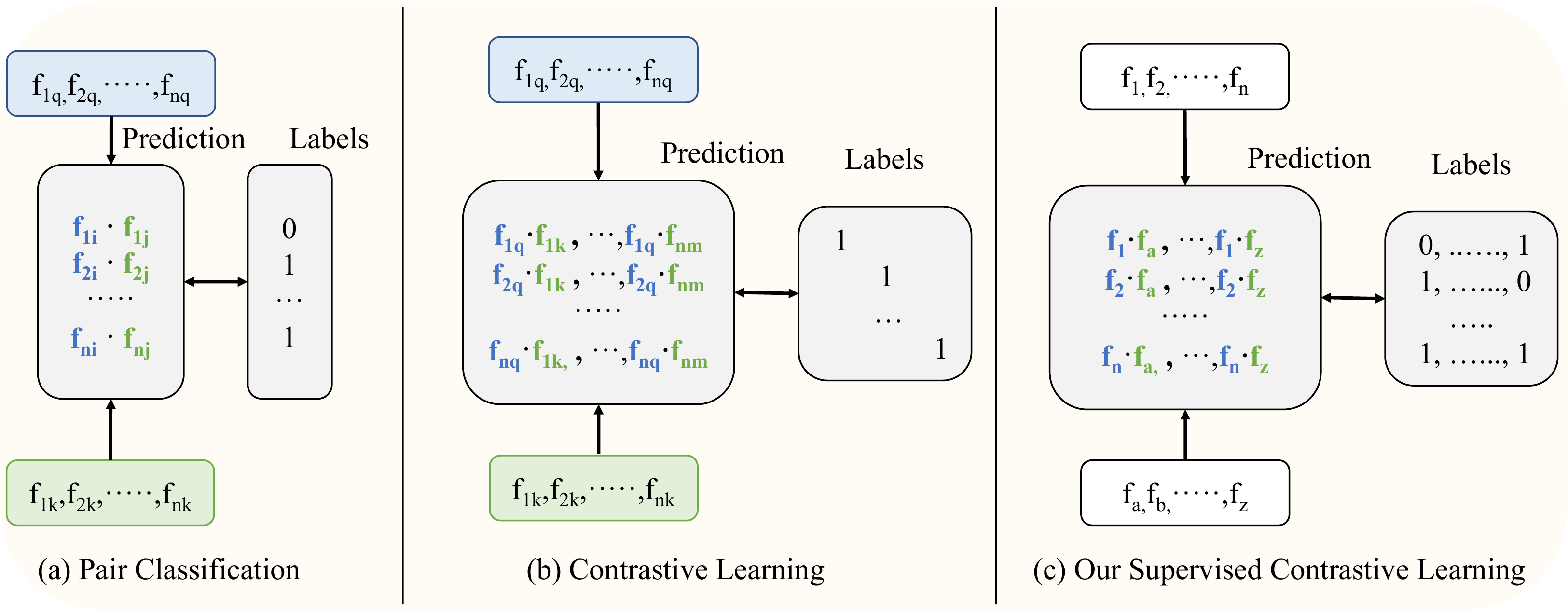}
	\caption{Different training paradigms based on the similar question dataset. Pair classification and contrastive learning needs pairwise inputting(query and key should be similar/positive), but our supervised contrastive learning could receive arbitrary instance inputting. Even a question do not have similar question, it can be input as dissimilar with other questions to learn the discrimination.}
	\label{fig:down}
\end{figure*}
\subsection{Mixed Unsupervised Contrastive Learning}
\par Our unsupervised pretraining was shown in Figure \ref{fig:pipeline} (step 1) as the first pretext task to push the text encoder to a suitable hidden space rather than directly do the multi-modal contrastive learning, as latter would easily cause the pattern collapse. 
\par After that, we can apply the multi-modal contrastive learning as shown in Figure \ref{fig:pipeline} (step 2): The Query/Key encoder have same architecture and initialization. Each network consist of three modules: visual encoder (ResNet-50), text encoder (BERT) and fusion module (multi-head attention). In this paper, we use the random masked words and clipped windows as the augmentation of questions as shown in Figure \ref{fig:pipeline} (b). The augment technologies of images are same as the \cite{Chen2020Improved}. Each input after augmentation consist of visual list $[\{I_1\},\{I_2\},..\{I_b\}]$ and text list(words and formula) $[T_1,T_2,...T_b]$. We will use the shuffle batch-normalization (BN) \cite{he2020momentum}, which collects the inputs from different GPUs to do the BN to avoid the device information leakage, to get key network features $k_1^+,k_2^+,..,k_b^+$. At the same time we have the $q_1,q_2,..,q_b$ positive features from the query network and $\{k_1^-\},\{k_2^-\},..,\{k_b^-\}$ negative features set from negative queues. Query network will be updated by the training loss (Eq.\ref{infonce}). 
\begin{equation}
    l_{q,k^+,\{k^-\}} = - log\frac{e^{q\bigcdot k^+ / \tau}}{e^{q\bigcdot k^+/ \tau}+\sum_{\{k^-\}} e^{q\bigcdot k^-/\tau}}
    \label{infonce}
\end{equation}
\par Key network parameters $p_{key}$ would be driven by a momentum update with the query encoder parameters $p_{query}$ follows Eq.\ref{moment}. 
\begin{equation}
    p_{key} = m* p_{key} + (1-m)* p_{query}
    \label{moment}
\end{equation}
where $m \in [0, 1)$ is a momentum coefficient (here we use m=0.999).
\par Note that, here we use the two-stage unsupervised contrastive learning mechanism to further enhance the representation ability of TQ-Net. In the first stage, MLM (Masked Language Model) was used to pre-train the text encoder (BERT in our model). In the second stage, \textbf{SEQ} (separately) and \textbf{UNI} (uniformly) pre-training were conducted. For SEQ, text modal and image modal were separately pre-trained in an unsupervised contrastive learning manner. For UNI, text modal and image modal were pre-trained in an unsupervised contrastive learning manner to view the question as an instance where augmentation applied to both text and image simultaneously (instance-level contrastive learning). Also, a comprehensive comparison of SEQ and UNI methods was shown in experiments.
\subsection{Supervised Contrastive Learning}
\par Though we have proposed an unsupervised learning method for heterogeneous question data, because the data structure could be greatly divergent, our model would sometimes over-fitting the noise. So based on our unsupervised pre-training method, we propose a supervised contrastive learning (SCL) fine-tuning method to build the knowledge anchor for models as shown in Figure \ref{fig:pipeline} (step 3). The positive pairs in SCL come from human-labelled similar questions rather than data augmentation. 
\par In educational applications, the similar question classification is an important task to evaluate the performance of educational model. And here we could adapt this binary classification dataset to do our supervised contrastive learning. 
\begin{equation}
    l_{scl} = - log\frac{e^{q_{seed}\bigcdot q_{sim} / \tau}}{e^{q_{seed}\bigcdot q_{sim}/ \tau}+\sum_{q_{o}\in \{q_{others}\}} e^{q\bigcdot q_{o}/\tau}}
    \label{scl}
\end{equation}
\par Let the seed question's feature be $q_{seed}$, and its similar question's feature is $q_{sim}$, others questions' features are $q_{others}$. The supervised contrastive learning is based on Eq.\ref{scl}, here the seed/similar question positive pair comes from human-labelled similar questions, and we use other questions in the dataset as the negative pairs of the seed. Note that, the label matrix no longer be diagonal, the label matrix was calculated by the ground truth: if questions pair is similar then label is 1, else the label is 0. Some questions could have multiple similar questions, so it's necessary to check the ground truth label matrix rather than applying augmentation as unsupervised contrastive learning used shown in Figure \ref{fig:down}. Another situation is even a question do not have similar questions, it can be input as dissimilar with other questions to learn the discrimination.
\section{Experiments}
\par The experiments of this paper mainly consist of two parts: (1) Similar test question, (2) Knowledge Point Prediction (Classification). In first part of experiments, we mainly explored the effective framework of TQ-Net to learning an effective representation of questions on both unsupervised and supervised ways. In second part, we evaluated TQ-Net compared to text-only methods and previous SOTA heterogeneous methods to illustrate the effectiveness of our proposal. 
\subsection{Dataset} The pre-training dataset we used came from an online education system and was collected from middle school math tests and exams as shown in Table \ref{tab:data}. About 30\% of questions contain images. The maximum number of images for one question is 10. We only use the text and images of questions and options without label to do the unsupervised representation learning.
\begin{table}[]
\renewcommand\tabcolsep{16pt}
    \begin{center}
    \scalebox{1.0}{
        \begin{tabular}{c|c|c}
        \toprule
             \multicolumn{3}{c}{Questions}\\
     \midrule
    Choice & Blank Filling & Calculation \\
    86753  & 42892 & 49225  \\ 
         \midrule
    \multicolumn{2}{c|}{Questions with images} & Images\\
     \multicolumn{2}{c|}{55635} & 91456\\
    \bottomrule
    \end{tabular}
    }
    \end{center}
\caption{Statistics of the questions dataset.}
\label{tab:data}
\end{table}
\subsection{Implementation details}
\par All our experiments use 8 1080-Ti GPUs. In the pre-training experiment, we follow the same settings of hyper-parameters with official MoCo v2 code \footnote{\url{https://github.com/facebookresearch/moco}}. In the similar test question classification task, we use SGD optimizer with lr=3e-4, momentum=0.9, and weight decay=5e-4. The training batch size was set to 32. In the knowledge classification task, we use the optimizer setting the same as similar test question, the training batch size was set to 64. Our augmentation for images are same as the visual contrastive learning methods \cite{he2020momentum,chen2020big}, such as resize, crop, colour jitter, flip and blur.
\subsection{Similar test question}
\par The similar test question is a relevance prediction task, which needs to predict whether two given questions are similar or not. In this task, training dataset contains 27637 question pairs, and 4526 pairs for evaluation. The labels were marked by multiple professionals to indicate whether the two questions in a pair are similar or not based on knowledge points. Specifically, each question in the train/test dataset will have at least 5 similar questions, and there is no overlap between the train and test datasets. In default, two random questions are not similar with a high probability. In our experimental setting, the question type is less considered, we hope that the model captures the knowledge/problem solving method/background description from the discrimination of similarity. 
\par We use precision at top-k as the metrics of performance. Specifically, we first build the similar mapping for each question according to the ground truth label, and each question has at least 5 similar questions on our designed testing datasets. Then based on the prediction result, we build the predicted mapping. For each question, we would pick the top-5 largest probability candidate from all the positive question predictions based on the cosine similarity of embeddings. Compared with the ground truth mapping, we calculated the precision at top-5 to illustrate the performance, as shown in Eq.\ref{cal_p}. In this paper, we choose the P@5 (responding to that each question will have at least 5 similar questions) as the main evaluation metrics of similar test question for the balance of contingency and generalization.
\begin{equation}
    P@k = \frac{1}{N} \sum_{i=0}^N \frac{card( S_{i-gt_k} \cap S_{i-pred_k}) }{card( S_{i-gt_k} )}
    \label{cal_p}
\end{equation}
Where, $S_{i-gt_k}$/$S_{i-pred_k}$ means the set of i-th sample top-k ground truth or predicted question index, $card$ means the number of elements in the set.
\par Similar test question is an essential question-based tasks for many educational applications. The performance of this task completely depends on the capability of the model to extract the representation of various questions. It's a difficult problem because of the heterogeneous, multi-modal data. Most solutions were only based on the text stream, and few studies as the pioneer to explore the heterogeneous solutions e.g. QuesNet \cite{yin2019quesnet}. In this section, we first explored to improve the representation learning of only text stream questions method, then introduced the image content to our framework.
\begin{table}[h]
\renewcommand\tabcolsep{16pt}
    \centering
    {
    \begin{tabular}{l|c|c|c}
\toprule
    Method & Pre. & Fin.  & P@5 \\
    \midrule
    $BERT_{base}$ & - & Pair  & 0.8164  \\ 
    $BERT_{\text{O}}$& MLM & Pair   & 0.8319 \\
    $BERT_{\text{E}}$& MLM & Pair & 0.8329 \\
    $BERT_{\text{E}}$ & CL & Pair & 0.8406 \\
    $BERT_{\text{E}}$& MCL & Pair& \textbf{0.8454}  \\
    \midrule
    $BERT_{base}$ & - & SCL   & 0.8242  \\ 
    $BERT_{\text{O}}$ & MLM & SCL & 0.8348 \\
    $BERT_{\text{E}}$ & MLM & SCL & 0.8386 \\
    $BERT_{\text{E}}$ & CL & SCL & 0.8396 \\
    $BERT_{\text{E}}$ & MCL & SCL & \textbf{0.8473} \\
    \bottomrule
    \end{tabular}
    }
    \caption{Experiments of only text stream with different settings. The subscript "base"/"O"/"E" respectively represent BERT pretrained settings are no pretrain/open-source/our educational dataset pretrained. "Pre." means pre-trained method. MLM means the mask language model pretrain, CL means contrastive learning, and MCL means our proposed mixed contrastive learning. "Fin." means fine-tune method. Pair/SCL method responding to Figure \ref{fig:down}. The best P@5 results are in \textbf{bold}.}
    \label{tab:text only}
\end{table}
\subsubsection{Text Only Input}
\par We started from the original text-only pair training method to illustrate that the contrastive learning method was more beneficial to learning the representation.  We only use the text stream of our TQ-Net, and ignore the images stream. We respectively used the masked language model (MLM), contrastive learning (CL) to do the pre-training, and also our proposed two-stage pre-training method MCL, which means the model first pre-training in MLM and then pre-training in CL framework (two-stage pre-training). To show the advantage of pre-training, we provided the performance of random initial and the official pre-training BERT by \cite{wolf-etal-2020-transformers}. 
\par According to the result in Table \ref{tab:text only}, compared to the MLM method, CL and MCL showed that contrastive learning was more helpful for representation learning. Note that, the pre-training task of CL was also a process to distinguish similarities. Thus, we want to figure out whether this method would help supervised representation learning. We design the SCL training method. Figure \ref{fig:down} shows the difference between SCL and traditional pair training on downstream similar question classification task. As the result shown in Table \ref{tab:text only}, it's reasonable to believe that SCL benefits the supervised representation learning and downstream tasks compared to the previous method. 
\subsubsection{Multi-modal Input}
\par
\begin{figure}[t]
	\begin{minipage}[t]{1\linewidth}
		\centering
		\includegraphics[width=0.8\textwidth, trim=0 0 0 0, clip]{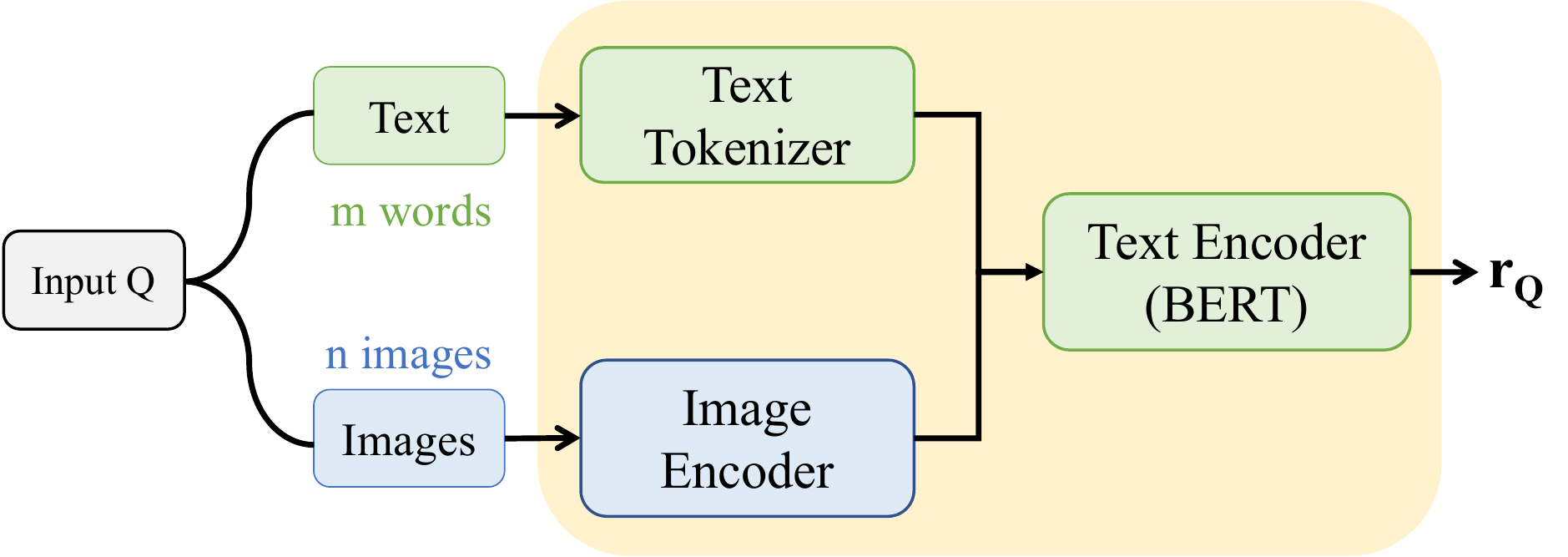}
		\centerline{(a) Joint representation}\medskip
	\end{minipage} 
	\begin{minipage}[t]{1\linewidth}
		\centering
		\includegraphics[width=0.8\textwidth, trim=0 0 0 0, clip]{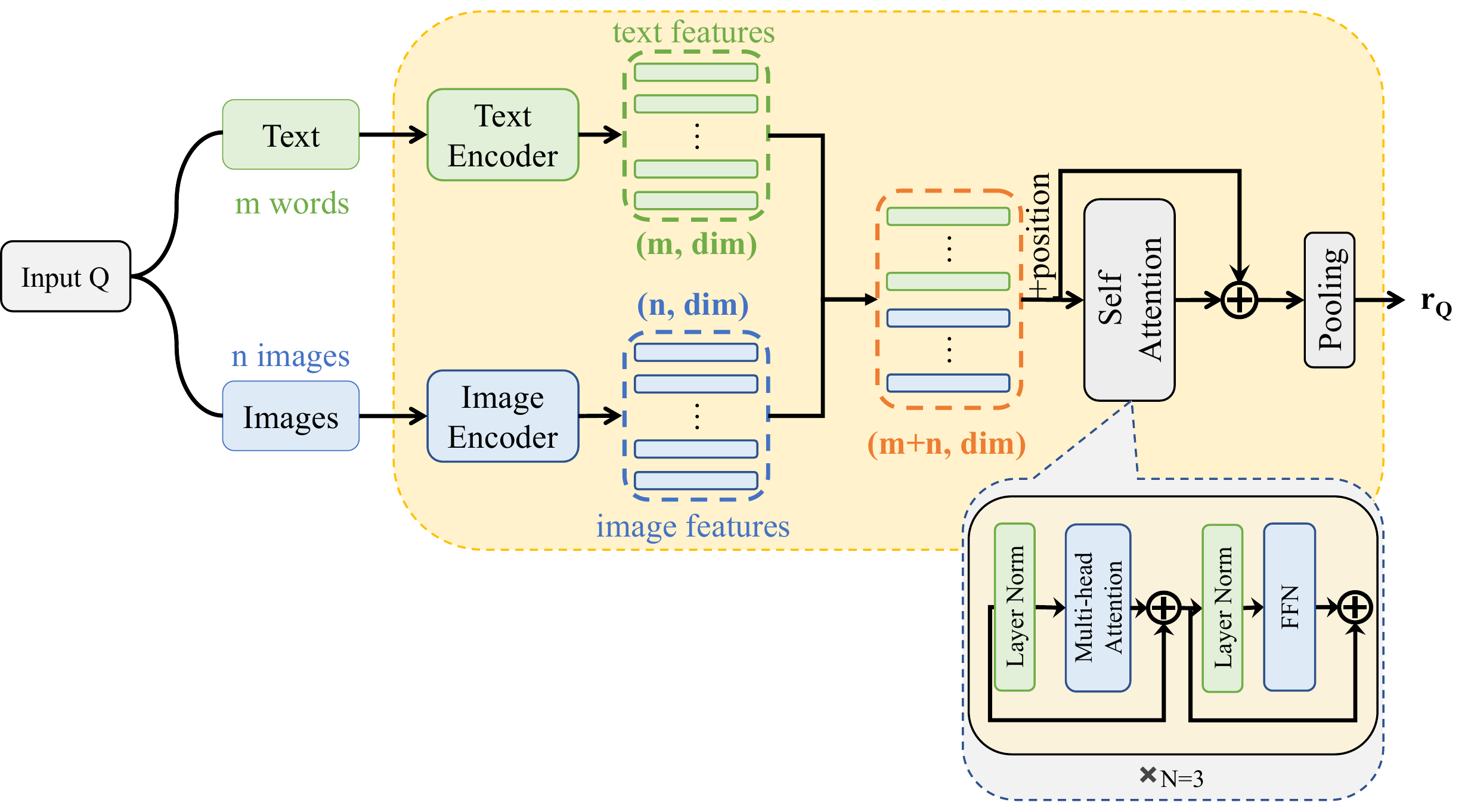}
		\centerline{(b) Coordinated representation}\medskip
	\end{minipage} 
	\caption{Different fusion methods of heterogeneous questions. Note that QuesNet (a) use the joint representation fusion, and TQ-Net (b) use coordinated representation fusion.}
	\label{fig:fuse}
\end{figure}

 Next, we would introduce the image content to improve the question representation learning. The main strategy of previous multi-modal fusion methods like \cite{sun2019videobert,li2020unicoder,radford2021learning,su2019vl,li2020unicoder} could be concluded into two types: joint or coordinated representation fusion as shown in Figure \ref{fig:fuse}. We first tested which one of these methods was more suitable for the question data representation learning.
\par The result as Table \ref{tab:fusion} shows that coordinated fusion may be suitable for question data. Thus, we use TQ-Net as a coordinated fusion method to process these data. We suppose these supplementary-each-other multi-modal data is hard to do the representation learning compared to image caption tasks. The reason may come from that the text is not the label or caption of the images, what's more, the relationship between the image and text modal tends to be more independent. 
\begin{table}
\renewcommand\tabcolsep{24pt}
    \centering
    \scalebox{0.90}{
    \begin{tabular}{l|c|c}
    \toprule
    Fusion & Fine-tune  & P@5 \\
    \midrule
    Joint & Pair &0.7969 \\
    Coordinated & Pair & \textbf{0.8329}\\
    \bottomrule
    \end{tabular}}
    \caption{The performance of different multi-modal fusion strategies. The best P@5 results are in \textbf{bold}.}
    \label{tab:fusion}
\end{table}
\par In view of that, we proposed two strategies to achieve a better unsupervised pre-training for heterogeneous question data to improve the downstream performance. (1) Separately pre-training the image/text encoders of TQ-Net. (2) Uniformly pre-training the TQ-Net as an instance encoder. 
\par To verify our argument, we test the contrastive way like CLIP, and the result as shown in Table \ref{tab:multimodal} is much worse (lower 6\%) than our instance contrastive learning. Our proposal greatly improves the capability of processing multi-modal test question data compared to the previous heterogeneous solution. Moreover, both SEP and UNI have better P@5 performance in pair and SCL mechanisms than text-only methods (Table \ref{tab:text only}). UNI was more beneficial to the unsupervised pre-training process than the frozen experiment. In the ablation study section, we would figure out where the performance improvement came from.
\begin{table}
\renewcommand\tabcolsep{20pt}
    \centering
    \scalebox{1}{
   \begin{tabular}{l|c|c}
 \toprule
    Pre-train & Fine-tune    & P@5 \\
    \midrule
    CLIP  & Pair  & 0.7995 \\ 
    QuesNet  &Pair & 0.8100\\
    \midrule
    \multirow{2}{*}{TQ-Net} 
    & Pair &0.8329 \\
     & SCL & \textbf{0.8531} \\
     % & SEP & MCL & Freeze & 0.7874 & 0.7134 & 0.7140 \\
     % & UNI & MCL & Freeze & 0.8841 & 0.8680 & 0.8377 \\
    \bottomrule
    \end{tabular}}
    \caption{Experiments of multi-modal input with different settings. The best P@5 results are in \textbf{bold}.}
    \label{tab:multimodal}
\end{table}

\subsection{Knowledge Point Prediction}
\par We also invite the knowledge point prediction task to evaluate our model knowledge anchor. We have 118919 questions on the training dataset, and 29565 on the testing dataset. Each question has knowledge point labels labelled by professional teachers. The total number of knowledge points is 391. So the knowledge point prediction task could be considered as a classification task, and we pick the F1 score as the metric of evaluation. We froze all the parameters of TQ-Net and introduce a linear classifier to do classification. For comparing, we provided the QuesNet and different initial experiments of BERT, both without freezing parameters.
\par According to the results Table \ref{tab:kp}, our TQ-Net shown great power to solve the unsupervised representation learning of heterogeneous test questions data.
\begin{table}[]
    \centering
    \renewcommand\tabcolsep{22pt}
    \scalebox{1}{
    \begin{tabular}{l|c}
    \toprule
       & {KP (class num =391)} \\
       \midrule
       Method   & F1\\
       \midrule
       QuesNet  &	0.557\\
       \midrule
       $BERT_{base}$ & 0.382 \\
       $BERT_{O}^\ast$ &  0.530 \\
       $BERT_{E}^\ast$ & 0.552 \\
       $TQ-Net_{MCL}^\ast$  & \textbf{0.602} \\
        \bottomrule
    \end{tabular}
    }
    \caption{Knowledge classification experiments. $\ast$ means the encoder parameters in experiments were frozen during the fine-tuning process. The best F1 results are in \textbf{bold}.}
    \label{tab:kp}
\end{table}

\begin{table}[]
    \renewcommand\tabcolsep{7pt}
    \centering
    \scalebox{1}{
    \begin{tabular}{c|c|c|c}
    \toprule
    Dataset& {Text-Text} & {Text-Image} & {Image-Image} \\
    \midrule
     Method  & \multicolumn{3}{c}{P@5} \\
     \midrule
     Text-only & 0.8688 & 0.7149 & 0.7882 \\
     Multi-modal  & \textbf{0.8866}  & \textbf{0.7194}  & \textbf{0.7960}\\
     \bottomrule
    \end{tabular}}
    \caption{Experiments on datasets consist of different pair structures. The best P@5 results are in \textbf{bold}.}
    \label{tab:ablation}
\end{table}

% \begin{figure}
%  \centering
%    \includegraphics[width=0.45\textwidth, trim=0 0 0 0, clip]{know.pdf}
% 	\caption{Each question has two different hierarchical levels of knowledge point labels, which called k2 and k3. The total number of k2 knowledge is 36, and k3 is 391. So the knowledge point prediction task could be considered as a multi-labels classification task.}
% 	\label{fig:know}
% \end{figure}

\subsection{Ablation Study}
\par To figure out where the improvement of the multi-modal performance came from, we respectively evaluated the model on different pair structure datasets in similar test question experiments: (1) Text-Text: two questions of pair do not contain images. (2) Text-Image: one question have images and the other one is text-only. (3) Image-Image: two questions of pairs both have images.
\par According to Table \ref{tab:ablation}, we found that the text input dominates the information of questions. Even without visual content, the model has the capability to represent the questions. However, the most improvement of TQ-Net performance came from the text-only question pairs. It shows that our multi-modal contrastive learning help to learn better representations of questions. Thus, we proposed that the visual content was necessary, which greatly benefits representation learning. 
\section{Conclusion}
\par In this paper, we improved the pre-training method of the previous text-only question representation with a two-stage contrastive learning mechanism. We also explored the fusion method of the image content, where a model named TQ-Net and instance-level contrastive learning was proposed to process the heterogeneous data. Moreover, we provided unsupervised and supervised training methods, which help the model to effectively learn the multi-modal representation of question data. We conducted extensive experiments to demonstrate the effectiveness of TQ-Net and improve the precision of downstream applications (especially relevance-related tasks). We will continue to explore possible improvements for the following in the future: (1) method for better fusion of the image content. (2) more effective contrastive learning framework for multi-modal or heterogeneous data. We hope this work builds a solid basis for future multi-modal and heterogeneous representation research and helps boost more educational applications.

\bibliography{custom}
\end{document}